\documentclass[letterpaper]{article} 
\usepackage{aaai2026}  
\nocopyright
\usepackage{times}  
\usepackage{helvet}  
\usepackage{courier}  
\usepackage[hyphens]{url}  
\usepackage{graphicx} 
\usepackage{natbib}  
\usepackage{caption} 
\usepackage{algorithm}
\usepackage{algorithmic}
\usepackage{placeins}
\usepackage{capt-of}
\usepackage{newfloat}
\usepackage{listings}
\DeclareCaptionStyle{ruled}{labelfont=normalfont,labelsep=colon,strut=off} 
\floatstyle{ruled}
\newfloat{listing}{tb}{lst}{}
\floatname{listing}{Listing}
\title{When Proxy Prediction Becomes Equation Reconstruction: Diagnostics and Residual Learning for Factor-Derived Proxy Supervision}

\author{
    Chayan Lahiri\textsuperscript{\rm 1}\equalcontrib,
    Ahmed Shafee\textsuperscript{\rm 2}\equalcontrib,
    Cody Fehringer\textsuperscript{\rm 3}
}

\affiliations{
    \textsuperscript{\rm 1}Department of Geosciences, Adams State University, Alamosa, Colorado, USA\\
    \textsuperscript{\rm 2}Department of Computer Science, Adams State University, Alamosa, Colorado, USA\\
    \textsuperscript{\rm 3}Devils Staircase Wildland Fire Module, Siuslaw National Forest, US Forest Service\\
    chayanlahiri@adams.edu, aashafee@adams.edu,  Cody.Fehringer@usda.gov
}

\usepackage{bibentry}

\begin{document}

\maketitle

\begin{abstract}
Scientific machine learning often relies on proxy targets computed from known domain factors when direct observations are limited. When those same factors are used as model inputs, however, high predictive accuracy may reflect reconstruction of the proxy-generating equation rather than robustness to degraded factor information. We study this problem in RUSLE-derived soil-loss proxy prediction under controlled degradation of the soil-erodibility factor \(K\). We introduce a diagnostic framework that combines degraded-formula references, classical tree-based baselines, matched direct and formula-feature predictors, contextual ablations, tail-error analysis, and degradation robustness scoring. We then propose RASPL, a formula-preserving residual framework that retains the degraded formula estimate as the prediction anchor and learns an adaptively gated contextual correction. RASPL substantially outperforms matched direct prediction and provides stronger degradation and tail robustness than treating the formula estimate as an ordinary input feature. Within RASPL, a compact statistical encoder achieves the highest macro-averaged \(R^2\) and lowest computational cost, whereas a convolutional encoder achieves the strongest degradation robustness and lowest Tail95 mean absolute error (MAE). These results establish formula preservation as the central design principle for robust learning from factor-derived proxy targets.
\end{abstract}

\section{Introduction}
\label{sec:Introduction}

In many scientific machine learning applications, direct measurements are sparse, expensive, or unavailable at scale. Watershed-scale soil-loss modeling is one such setting: field measurements are valuable for physical validation, but spatially complete erosion labels are rarely available for large-scale supervised learning \cite{Benavidez2018RUSLEReview,Borrelli2017GlobalErosion}. Consequently, soil-erosion assessments often rely on Revised Universal Soil Loss Equation (RUSLE)-derived proxy estimates constructed from rainfall erosivity, soil erodibility, topographic, and cover-management factors \cite{Panagos2015,Borrelli2017GlobalErosion,Li2023RUSLE}. When these same factors are used as model inputs, however, high predictive accuracy may reflect reconstruction of the proxy-generating equation rather than robustness to degraded, missing, or uncertain factor information.


We study this setting as \emph{factor-derived proxy supervision}, where the target is computed from known scientific factors rather than independently observed outcomes. Using RUSLE-derived soil-loss proxy prediction as a case study \cite{Ge2023,land13020174,w17091351}, the target is
\begin{equation}
A = R \times K \times LS \times C \times P,
\label{eq}
\end{equation}
where \(R\), \(K\), \(LS\), \(C\), and \(P\) denote rainfall erosivity, soil erodibility, topographic, cover-management, and support-practice factors, respectively. Following prior work, we set \(P=1\) in the study region~\cite{Ahmet_Mediterranean}; therefore, the implemented target reduces to \(A=R\times K\times LS\times C\). When the four varying factors are provided as inputs, the target can be reconstructed directly from the known equation. Strong learned-model performance under this setting may therefore reflect approximation of the proxy-generating equation rather than robustness when factor information is degraded.

To separate reconstruction from robust learning, we evaluate models under complete, noisy, coarsened, and masked factor regimes, together with a separate missing-factor stress test. The evaluation combines matched degraded-formula references, classical center and neighborhood baselines, matched direct and formula-feature predictors, contextual-representation ablations, tail-error analysis, and degradation robustness scoring. These diagnostics shift evaluation from reproducing the proxy under complete information to remaining robust when an important factor is degraded.

We further introduce RASPL, a formula-preserving residual framework that retains the degraded proxy-formula estimate as the prediction anchor and learns an adaptively gated contextual correction. The framework supports correction from either compact neighborhood statistics or learned convolutional representations, enabling a controlled analysis of whether useful information arises from neighborhood summaries, raw local values, or their spatial arrangement. 
Our contributions are as follows:
\begin{itemize}
\item We formalize \emph{equation reconstruction} as a failure mode in factor-derived proxy supervision, where high predictive accuracy may not imply robustness to degraded factor information.
\item We introduce a diagnostic protocol based on controlled factor degradation, matched formula references, classical and neural baselines, contextual-representation ablations, tail metrics, and degradation robustness scoring.
\item We propose RASPL, a formula-preserving residual framework that retains the degraded proxy-formula estimate as the prediction anchor and learns an adaptively gated contextual correction. Controlled statistical and convolutional evaluations demonstrate the benefit of formula preservation and characterize the tradeoff among average accuracy, tail robustness, and computational complexity.
\end{itemize}

\section{Related Work}
\label{sec:related_work}

\textbf{Factor-derived proxy targets and machine-learning-enhanced erosion modeling.} Empirical formulas are widely used to construct environmental proxy targets when direct measurements are sparse or unavailable. In soil-loss modeling, RUSLE derives erosion proxies from rainfall erosivity, soil erodibility, topographic, and cover-management factors \cite{Benavidez2018RUSLEReview,Panagos2015,Li2023RUSLE}. Prior work combines RUSLE-derived factors, remote-sensing data, and machine learning to estimate soil loss, improve factor layers such as the \(K\) factor, refine erosion maps, and analyze erosion drivers \cite{land13020174,Zeghmar2024, Abiye2025SoilErodibility,Ge2023,w17091351}. Related studies use environmental predictors to map erosion susceptibility from categorical labels or regional erosion inventories \cite{BOUAMRANE2024998,Gelete_2024,MomeniDamaneh2023ErosionML,Tkeshelashvili2024Tsageri,Olii2025SaddangSEV,Priyadharshini2025GeomorphicRisk,Islam2025GullyErosionML}. In this literature, RUSLE generally remains a fixed empirical formulation while machine learning improves its inputs, outputs, or downstream analysis. Our work instead treats the RUSLE-derived quantity itself as the supervision target and evaluates controlled degradation of its generating factors to distinguish proxy-equation reconstruction from degraded-factor robustness and to assess whether formula-preserving corrections remain effective.

\textbf{Spatial deep learning, foundation models, and formula-aware prediction.}
Prior studies have used CNNs, Transformers, and geospatial representation-learning methods to capture local context and multichannel spatial structure \cite{s23218966,sentinelData}. More recent geospatial foundation models, including SatMAE, CROMA, and GeoCLIP, learn transferable representations from large-scale Earth-observation data \cite{Cong2022SatMAE,CROMA,GeoCLIP}. These methods primarily improve feature representation and transfer across geospatial tasks, whereas our work addresses the distinct setting in which the supervision target is generated by a known scientific equation. Related physics-informed and structure-aware methods incorporate scientific knowledge through model architectures, constraints, or training objectives \cite{raissi2019physics,karniadakis2021physics,willard2022integrating}. In our setting, the known equation directly defines the supervision target and is retained as the prediction anchor rather than serving only as an auxiliary constraint. We further compare statistical and convolutional contextual encoders to determine whether predictive gains arise from neighborhood summaries, raw local values, or their spatial arrangement.

\textbf{Robustness under degraded features.}
Prior work addresses missing, noisy, or corrupted inputs through imputation, augmentation, robust losses, and corruption-based evaluation \cite{yoon2018gain,shorten2019survey,huber1964robust,hendrycks2019benchmarking}. We instead use controlled degradation as both a robustness evaluation and a diagnostic for proxy supervision. We compare complete, noisy, coarsened, and masked factor regimes, together with a separate missing-factor stress test, to distinguish equation reconstruction from degraded-factor robustness.

\section{Problem Setup and Diagnostics}
\label{sec:setup}

\subsection{Proxy Supervision and Equation Reconstruction}

We study supervised prediction of a \emph{proxy} target constructed from known scientific factors rather than from independent field measurements. For sample \(i\),
\begin{equation}
A_i = g(f_{i,1},\ldots,f_{i,m}),
\end{equation}
where \(g(\cdot)\) is a fixed generating function and \(f_{i,j}\) denotes the value of factor \(j\) at sample \(i\). In our RUSLE setting,
\begin{equation}
A_i = R_i \times K_i \times LS_i \times C_i,
\end{equation}
where \(A_i\) is taken from the precomputed soil-loss raster.

When the same factors used to define \(A_i\) are also provided as model inputs, high accuracy may reflect \emph{complete-factor proxy reconstruction}---that is, approximation of \(g(\cdot)\) itself---rather than robust use of contextual information under imperfect factors. We therefore evaluate each predictor against a \textbf{formula reference}
\begin{equation}
\hat{A}_{\mathrm{formula},i}
=
R_i \times \tilde{K}_i \times LS_i \times C_i,
\end{equation}
where \(\tilde{K}_i\) is the degraded center-pixel value, or a fixed training-set fallback when \(K\) is absent as described in the following subsection. Under \(K_{\mathrm{full}}\), this reference nearly recovers \(A_i\) and serves as a reconstruction benchmark. Under degraded regimes, comparison with the corresponding degraded formula reference isolates gains beyond direct equation evaluation. We do not assess validity against independently observed soil-loss measurements.

\subsection{Degraded-Factor Regimes}
\label{subsec:degradation}

We perturb only the erodibility factor \(K\), denoting by \(\tilde{K}\) the version available to both the predictors and the formula reference. The main matched evaluation uses five regimes: \(K_{\mathrm{full}}\), \(K_{\mathrm{noise\_020}}\), \(K_{\mathrm{coarse\_8}}\), \(K_{\mathrm{mask\_050}}\), and \(K_{\mathrm{center\_mask\_050}}\). Complete absence of \(K\) is evaluated separately through \(K_{\mathrm{missing}}\).

\textbf{Raster-level regimes.} For \(K_{\mathrm{full}}\), \(\tilde{K}=K\). 
For \(K_{\mathrm{noise\_020}}\), \(\tilde{K}=K\exp(\epsilon)\), where \(\epsilon\sim\mathcal{N}(0,0.20^2)\).
For \(K_{\mathrm{coarse\_8}}\), we apply nodata-aware \(8\times8\) block averaging followed by piecewise-constant upsampling. Invalid pixels are excluded from the block means, and the original nodata mask is restored. Noise and coarsening are applied to the complete \(K\) raster before window extraction.

\textbf{Sample-level masking.} Let \(\bar{K}_{\mathrm{train}}\) denote the mean center-pixel \(K\) value over the training samples. In \(K_{\mathrm{mask\_050}}\), approximately \(50\%\) of the samples in each split are masked independently. For each masked sample, the complete \(K\) window and the center value used by the formula reference are replaced by \(\bar{K}_{\mathrm{train}}\). In \(K_{\mathrm{center\_mask\_050}}\), the same masking rate is used, but only the center pixel and the corresponding formula-reference value are replaced; the neighboring \(K\) values remain available. Unmasked samples retain their original \(K\) values. The reliability vector \(\mathbf{z}\) contains a mask or missingness indicator, a one-hot degradation tag, seven \(K\)-neighborhood statistics when \(K\) is observed, and the absolute center-to-neighborhood-mean difference
\(\left|\tilde{K}_{\mathrm{center}}-\tilde{K}_{\mathrm{mean}}\right|\).
All \(K\)-derived entries are set to zero under \(K_{\mathrm{missing}}\).
RASPL-MLP-STATS and RASPL-CNN-RAW+STATS use the full reliability vector in both the gating and residual branches.
For RASPL-MLP-CENTER, the gate conditions on the full reliability vector, whereas the residual branch combines the four center-pixel channels with a reduced eight-dimensional reliability subset consisting of the mask or missingness indicator and the one-hot degradation tag.
The residual branches of RASPL-MLP-FLAT and RASPL-CNN-RAW likewise use a reduced eight-dimensional non-statistical subset of the reliability vector together with their respective flattened-window and convolutional representations.

\textbf{Missing-\(K\) stress test.} Under \(K_{\mathrm{missing}}\), observed \(K\) is omitted from the spatial window, which becomes \([R,LS,C]\) together with a constant missingness-indicator plane. The formula reference is
\begin{equation}
\hat{A}_{\mathrm{formula},i}
=
R_i
\times
\bar{K}_{\mathrm{train}}
\times
LS_i
\times
C_i.
\end{equation}
Results for \(K_{\mathrm{missing}}\) are reported separately and excluded from the main degradation averages and rankings.

\subsection{Diagnostic Criteria}
\label{subsec:diagnostics}

We use five diagnostics. \textbf{Formula reconstruction} tests whether strong performance under \(K_{\mathrm{full}}\) is explained by direct recovery of the generating equation. \textbf{Formula preservation} (Q1) compares matched direct, formula-feature, and RASPL predictors using statistical MLP and convolutional CNN encoders. \textbf{Spatial-context gain} (Q2) compares center-only, neighborhood-statistics, flattened-window, and convolutional-window representations. \textbf{Tail robustness} is evaluated using Tail95 MAE and the Tail95 underprediction rate over positive samples whose target values are at or above the \(95\)th percentile. Underprediction is defined as $\hat{A} < 0.8A.$
\textbf{Degradation robustness} is summarized across regimes using the degradation robustness score (DRS):
\begin{equation}
\begin{array}{rl}
\mathrm{DRS}={}&
0.35\,\widetilde{R^2_{\mathrm{all}}}
+
0.25\,\widetilde{R^2_{\mathrm{pos}}}
\\[2pt]
&
+
0.25\,\widetilde{\mathrm{T95MAE}}
+
0.15\,\widetilde{\mathrm{T95Under}}.
\end{array}
\end{equation}
Each component is min--max normalized within the stated comparison pool, separately for each regime across models. The tail-error components are direction-reversed so that higher values are consistently better. The component metrics are also reported separately. For RASPL, stability is evaluated using the effective correction \(\left|\alpha\delta\right|\), rather than the raw residual \(\delta\) alone as defined in the RASPL Method section.

\section{RASPL Method}
\label{sec:raspl}

\subsection{Motivation}

Under complete factors, strong proxy-prediction performance may reflect reconstruction of the generating equation rather than meaningful use of contextual information. Under degraded factors, however, accurate prediction requires compensation for corrupted or missing inputs. Direct predictors can use contextual information, but they treat the proxy as an ordinary regression target and do not preserve the known formula structure. RASPL instead retains the available formula estimate as the prediction anchor and learns when contextual and reliability information should correct it. To isolate the role of contextual representation, we compare statistical and convolutional residual encoders. MLP-STATS tests whether neighborhood summary statistics are sufficient for correction, whereas CNN-RAW+STATS tests whether the spatial arrangement of local values provides additional information. Center-only, flattened-window, and raw-window variants serve as controlled ablations.

\begin{figure*}[t]
\centering
\includegraphics[width=\textwidth]{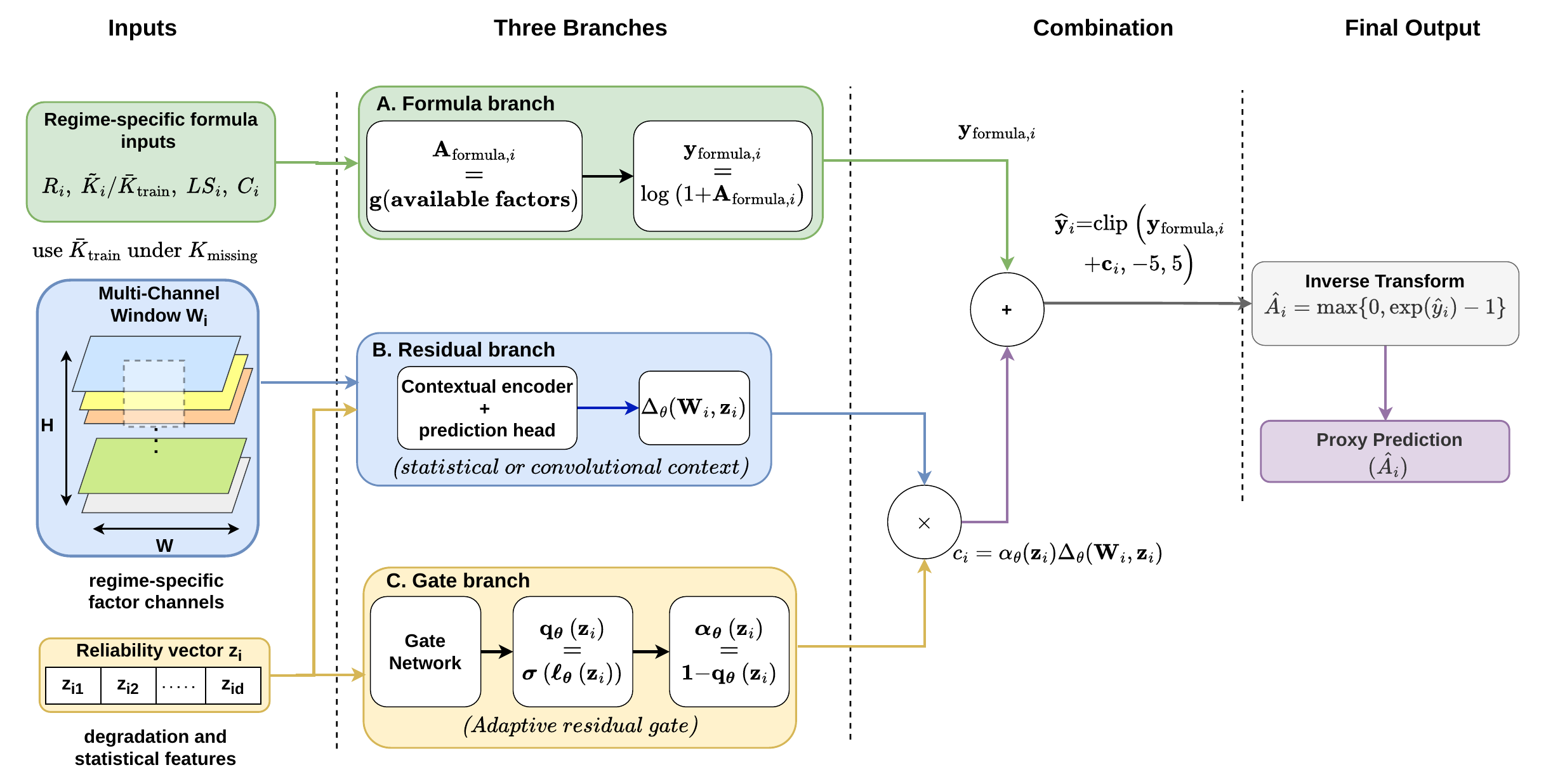}
\caption{Overview of RASPL. The degraded formula estimate is retained as the prediction anchor, and an adaptively gated contextual residual is added in log space before the prediction is mapped back to the proxy scale.}
\label{fig:raspl_architecture}
\end{figure*}

\subsection{Formula-Preserving Residual Prediction}

As shown in Figure~\ref{fig:raspl_architecture}, RASPL combines a fixed formula estimate, a contextual residual encoder, and an adaptive gate. Let \(W_i\) denote the regime-specific local input and \(\mathbf{z}_i\) the reliability vector defined in the Degraded-Factor Regimes subsection. The formula estimate \(A_{\mathrm{formula},i}\) uses the degraded center-pixel value \(\tilde{K}_i\) when \(K\) is observed and the training-mean fallback \(\bar{K}_{\mathrm{train}}\) under \(K_{\mathrm{missing}}\).

Because the proxy is nonnegative and skewed, prediction is performed in log space:
\[
y_i=\log(1+A_i),
\qquad
y_{\mathrm{formula},i}=\log(1+A_{\mathrm{formula},i}).
\]
The model produces a raw residual proposal \(\Delta_{\theta}(W_i,\mathbf{z}_i)\) and a gate value
\[
q_{\theta}(\mathbf{z}_i)=\sigma\!\left(\ell_{\theta}(\mathbf{z}_i)\right),
\qquad
\alpha_{\theta}(\mathbf{z}_i)=1-q_{\theta}(\mathbf{z}_i),
\]
where \(\ell_{\theta}\) is a learned scalar logit computed from the reliability vector. The predicted log value and proxy-scale prediction are
\[
\begin{array}{rcl}
\hat{y}_i
&=&
\mathrm{clip}\!\left(
y_{\mathrm{formula},i}
+
\alpha_{\theta}(\mathbf{z}_i)\Delta_{\theta}(W_i,\mathbf{z}_i),
-5,5
\right),\\[4pt]
\hat{A}_i
&=&
\max\left\{0,\exp(\hat{y}_i)-1\right\}.
\end{array}
\]
The effective correction applied to the formula estimate is therefore $alpha_{\theta}(\mathbf{z}_i)\Delta_{\theta}(W_i,\mathbf{z}_i),$
rather than the raw residual alone. A small value of \(\alpha_{\theta}\) retains the formula estimate, whereas a large value permits a stronger contextual correction. The gate is learned end-to-end without reliability labels and is interpreted as a residual-allocation weight rather than calibrated uncertainty.

\subsection{Contextual Residual Encoders}

MLP-STATS uses the full reliability vector \(\mathbf{z}_i\), including neighborhood summary statistics, in both the residual and gating branches. This encoder tests whether compact statistical context is sufficient to correct the degraded formula estimate.

CNN-RAW+STATS applies a compact two-layer CNN to the raw local input \(W_i\), pools the resulting representation, concatenates it with the full reliability vector \(\mathbf{z}_i\), and predicts the residual proposal \(\Delta_{\theta}\). Its gate value \(q_{\theta}\) also conditions on the full reliability vector. This encoder tests whether the spatial arrangement of local values provides information beyond neighborhood summaries.

The remaining variants use the same formula-preserving prediction rule with different contextual inputs. MLP-CENTER uses the four center-pixel channels in the residual branch; its gate conditions on the full reliability vector, whereas its residual branch uses a reduced eight-dimensional non-statistical reliability subset. MLP-FLAT and CNN-RAW use flattened-window and convolutional raw-window encoders, respectively, together with reduced eight-dimensional non-statistical reliability inputs rather than the full reliability vector.

\subsection{Training Objective}

All four channels of \(W\) are standardized using statistics from the training split. In \(\mathbf{z}\), only the continuous \(K\)-derived entries are standardized; the indicator and one-hot degradation entries are left unchanged. RASPL is trained against the unstandardized log targets \(y_i\) and formula estimates \(y_{\mathrm{formula},i}\). The matched direct baselines instead predict a training-standardized log target, while the formula-feature baseline receives a training-standardized version of \(\log(1+A_{\mathrm{formula}})\) as an additional scalar input. RASPL minimizes a weighted log-space regression loss with 
\[
w_i
=
1
+
\lambda_{\mathrm{pos}}\mathbf{1}[A_i>0]
+
\lambda_{\mathrm{tail}}\mathbf{1}[A_i\geq Q_{95}^{\mathrm{train}}],
\]
where \(Q_{95}^{\mathrm{train}}\) is the \(95\)th percentile of the positive training targets when more than ten positive samples are available and the \(95\)th percentile of all training targets otherwise. We set \(\lambda_{\mathrm{pos}}=0.5\) and \(\lambda_{\mathrm{tail}}=1.0\). The main loss is
\[
\mathcal{L}_{\mathrm{main}}
=
\frac{1}{N}
\sum_{i=1}^{N}
w_i(\hat{y}_i-y_i)^2.
\]

A residual penalty discourages large raw corrections when the gate favors retention of the formula estimate:
\[
\mathcal{L}_{\mathrm{res}}
=
\frac{1}{N}
\sum_{i=1}^{N}
q_{\theta}(\mathbf{z}_i)\,
\Delta_{\theta}(W_i,\mathbf{z}_i)^2.
\]
The total objective is $\mathcal{L}
=
\mathcal{L}_{\mathrm{main}}
+
\lambda_{\mathrm{res}}\mathcal{L}_{\mathrm{res}},$ with \(\lambda_{\mathrm{res}}=0.01\).

\subsection{Matched Direct and Formula-Feature Baselines}

The direct baselines remove the formula anchor, adaptive gate, and residual addition while retaining the same contextual encoders and regime-matched inputs as their corresponding RASPL variants. They are trained to predict the training-standardized log target directly, and inverse standardization maps their outputs back to the proxy scale.

The matched formula-feature baseline additionally receives the training-standardized log formula estimate as an ordinary input feature but still predicts the complete target directly. It does not preserve the formula estimate through an explicit residual connection. Comparing these baselines with RASPL isolates the effect of retaining the proxy-generating equation as the prediction anchor rather than ignoring it or treating it as an ordinary input feature.

\section{Experimental Setup}

\textbf{Study Setting and Proxy Target.}
We evaluate the framework on the RUSLE-derived proxy task defined in the Problem Setup and Diagnostics section. Each sample contains rainfall erosivity \(R\), soil erodibility \(K\), topographic factor \(LS\), and cover-management factor \(C\), derived from PRISM precipitation~\cite{daly2002prism}, USDA SSURGO~\cite{usda_ssurgo}, USGS 3DEP DEM~\cite{usgs_3dep}, and Sentinel-2 LULC~\cite{sentinelData}, respectively. The target \(A\) is therefore a factor-derived proxy rather than an independently measured soil-loss observation. The main experiments use co-registered \(3\times3\) factor windows, center-level features, and the reliability vector \(\mathbf{z}_i\) defined in the Degraded-Factor Regimes subsection; the window-size study replaces the spatial input with \(5\times5\) and \(7\times7\) windows while holding the remaining protocol fixed.

\textbf{Evaluation Regimes.}
All matched comparisons use seeds \(\{42,123,999\}\). The formula-preservation comparison (Q1) includes
\(K_{\mathrm{full}}\),
\(K_{\mathrm{noise\_020}}\),
\(K_{\mathrm{coarse\_8}}\),
\(K_{\mathrm{mask\_050}}\), and
\(K_{\mathrm{center\_mask\_050}}\),
with \(K_{\mathrm{full}}\) also serving as the reconstruction diagnostic. The contextual-encoder comparison (Q2) uses
\(K_{\mathrm{noise\_020}}\),
\(K_{\mathrm{coarse\_8}}\), and
\(K_{\mathrm{center\_mask\_050}}\).
The \(K_{\mathrm{missing}}\) regime is evaluated separately as the strongest stress test and is excluded from the shared ranking pools.

\subsection{Models and Baselines}

\textbf{Formula and Tree-Based References.}
Formula references compute \(A_{\mathrm{formula}}\) using the regime-specific available factors defined in the Degraded-Factor Regimes subsection; under \(K_{\mathrm{missing}}\), they use the training-mean fallback \(\bar{K}_{\mathrm{train}}\). Random Forest and XGBoost references use either center-level factors or engineered neighborhood statistics comprising the center value, mean, standard deviation, minimum, maximum, median, lower and upper quartiles, range, and local gradient features. A two-stage XGBoost model is additionally included in the complete-factor reconstruction diagnostic.

\textbf{Matched Formula-Preservation Models.}
The matched CNN comparison evaluates direct prediction without the formula estimate, direct prediction with the standardized log formula estimate as an ordinary feature, and RASPL with the formula estimate retained as an adaptively gated residual anchor. The same three-way comparison is applied to statistics-only MLP encoders over the Q2 regime subset. Additional RASPL ablations vary only the contextual representation supplied to the residual and gating components.

\textbf{Window-Size Variants.} The convolutional window-size study compares RASPL-CNN-RAW+STATS with \(3\times3\), \(5\times5\), and \(7\times7\) windows over the Q2 subset while holding the encoder family, channel definitions, seeds, optimization procedure, and checkpoint-selection rule fixed.

\textbf{Training Protocol.} Neural models follow the transformations and objectives defined in the RASPL Method section. RASPL uses the weighted residual objective on unstandardized log targets and formula estimates, whereas direct and formula-feature baselines use weighted mean-squared error on standardized log targets. Formula-feature models additionally receive the standardized log formula estimate, and all outputs are inverse-transformed to the proxy scale.

All neural models use AdamW with batch size \(512\), weight decay \(10^{-4}\), at most \(50\) epochs, patience \(5\), gradient clipping at \(1.0\), and learning-rate selection from \(\{10^{-4},3\times10^{-4},10^{-3}\}\). Checkpoints are selected using the lowest finite validation loss.

Tree-based and neural models use identical data partitions. Matched aggregates include only shared, completed regime--seed cells.

\subsection{Evaluation Metrics}
\label{subsec:evaluation_metrics}

We report \(R^2\), MAE, and RMSE over all test samples and over the positive-target subset \(A_i>0\), denoted by the subscripts \(\mathrm{all}\) and \(\mathrm{pos}\), respectively.

Let
\[
\mathcal{T}_{95}
=
\left\{
i:A_i\geq Q_{95}^{\mathrm{train}}
\right\},
\]
where \(Q_{95}^{\mathrm{train}}\) is the \(95\)th percentile of the positive training targets when more than ten positive samples are available and the \(95\)th percentile of all training targets otherwise. The tail metrics are
\[
\mathrm{Tail95\ MAE}
=
\frac{1}{|\mathcal{T}_{95}|}
\sum_{i\in\mathcal{T}_{95}}
|A_i-\hat{A}_i|
\]
and
\[
\mathrm{Tail95\ Under}
=
\frac{1}{|\mathcal{T}_{95}|}
\sum_{i\in\mathcal{T}_{95}}
\mathbf{1}
\left[
\hat{A}_i<0.8A_i
\right].
\]

The degradation robustness score combines normalized
\(R^2_{\mathrm{all}}\),
\(R^2_{\mathrm{pos}}\),
Tail95 MAE, and Tail95 underprediction using the weights defined in the Diagnostic Criteria subsection. Normalization is performed separately within each regime and comparison pool, with the tail-error components direction-reversed so that higher DRS is better. Consequently, DRS values from different normalization pools are not directly comparable.


\section{Results and Analysis}



\textbf{Complete-Factor Reconstruction and Degradation Sensitivity.} Table~\ref{tab:full_factor_reconstruction} shows that the formula reference attains exactly \(R^{2}_{\mathrm{all}}=1.0000\) under \(K_{\mathrm{full}}\), confirming that the proxy target is exactly reconstructable from the generating factors when they are fully available. The matched tree baselines likewise achieve strong complete-factor performance, with three-seed mean \(R^{2}_{\mathrm{all}}\) ranging from \(0.8387\) for XGB-center to \(0.9471\) for two-stage XGBoost. Mild multiplicative noise in \(K\) produces only small reductions: the formula reference remains at \(0.9560\), while RF-center, XGB-center, and two-stage XGBoost attain \(0.8794\), \(0.8312\), and \(0.9252\), respectively. Spatial coarsening reduces \(R^{2}_{\mathrm{all}}\) for the formula reference to \(0.9259\) and for all three learned baselines to values between \(0.7535\) and \(0.8250\).

Broad masking produces the largest deterioration. The formula reference decreases to a three-seed mean of \(0.5114\), while RF-center, XGB-center, and two-stage XGBoost decrease to \(0.4622\), \(0.3256\), and \(0.4273\), respectively. RF-center, XGB-center, and two-stage XGBoost use center-level factor information and therefore receive the same effective center \(K\) input under \(K_{\mathrm{mask}}\) and \(K_{\mathrm{center\mbox{-}mask}}\), yielding identical results across the two regimes. The formula reference, which is also computed from center-pixel factors, exhibits the same behavior. Together, these results demonstrate that strong complete-factor accuracy can reflect exploitation of the deterministic proxy-generating relationship and does not, by itself, establish robustness to degraded factor information.


\begin{table}[t]
\centering
\small
\setlength{\tabcolsep}{2.2pt}
\renewcommand{\arraystretch}{1.08}
\caption{Complete-factor reconstruction and \(K\)-degradation sensitivity. Values are mean \(R^{2}_{\mathrm{all}}\) over seeds \(\{42,123,999\}\). Columns correspond to \(K_{\mathrm{full}}\), \(K_{\mathrm{noise}}\), \(K_{\mathrm{coarse}}\), \(K_{\mathrm{mask}}\), and \(K_{\mathrm{center\mbox{-}mask}}\), respectively.}
\label{tab:full_factor_reconstruction}
\begin{tabular}{@{}lccccc@{}}
\hline
Model & Full & Noise & Coarse & Mask & C-mask \\
\hline
Formula ref. & 1.0000 & 0.9560 & 0.9259 & 0.5114 & 0.5114 \\
RF-center & 0.8895 & 0.8794 & 0.7595 & 0.4622 & 0.4622 \\
XGB-center & 0.8387 & 0.8312 & 0.7535 & 0.3256 & 0.3256 \\
Two-stage XGB & 0.9471 & 0.9252 & 0.8250 & 0.4273 & 0.4273 \\
\hline
\end{tabular}
\end{table}


\textbf{Formula Preservation.} We first isolate the effect of formula preservation using matched statistical encoders evaluated over
\(K_{\mathrm{noise\_020}}\),
\(K_{\mathrm{coarse\_8}}\), and
\(K_{\mathrm{center\_mask\_050}}\)
with seeds \(\{42,123,999\}\). The three models use the same neighborhood-statistics representation but differ in how they incorporate the degraded formula estimate: the direct model does not receive it, the formula-feature model treats it as an ordinary input feature, and RASPL retains it as the prediction anchor.

\begin{table}[t]
\centering
\small
\renewcommand{\arraystretch}{1.12}
\setlength{\tabcolsep}{3.5pt}
\caption{Matched statistical-encoder comparison over three degraded-factor regimes and three seeds. DRS is normalized within this focused three-model comparison pool.}
\label{tab:main_results}
\begin{tabular}{lccc}
\hline
Model &
\(R^2_{\mathrm{all}}\) \(\uparrow\) &
T95 MAE \(\downarrow\) &
DRS \(\uparrow\) \\
\hline
RASPL-MLP-STATS
& \textbf{0.8343}
& \textbf{0.2385}
& \textbf{0.9986} \\
MLP-STATS+formula
& 0.8188
& 0.2994
& 0.9533 \\
Direct-MLP-STATS
& 0.1017
& 1.1573
& 0.0000 \\
\hline
\end{tabular}
\end{table}

Table~\ref{tab:main_results} shows that neighborhood statistics alone are insufficient for reliable prediction in this matched MLP setting. Adding the degraded formula estimate as an input feature increases macro \(R^2_{\mathrm{all}}\) from \(0.1017\) to \(0.8188\) and reduces Tail95 MAE from \(1.1573\) to \(0.2994\), indicating that access to the formula estimate accounts for most of the recovery in average accuracy. Explicitly preserving the formula estimate through RASPL provides a further increase in \(R^2_{\mathrm{all}}\) to \(0.8343\) and reduces Tail95 MAE to \(0.2385\).

Across the nine matched regime--seed cells, RASPL-MLP-STATS exceeds Direct-MLP-STATS in \(R^2_{\mathrm{all}}\) and Tail95 MAE in all nine cells, with mean improvements of \(0.733\) in \(R^2_{\mathrm{all}}\) and \(0.919\) in Tail95 MAE. Relative to the formula-feature model, RASPL improves both metrics in eight of nine cells, with mean improvements of \(0.015\) in \(R^2_{\mathrm{all}}\) and \(0.061\) in Tail95 MAE. Tail95 underprediction is more mixed, with RASPL improving five of nine cells, indicating that tail-error magnitude and underprediction frequency capture related but distinct behavior.

\begin{table}[t]
\centering
\small
\renewcommand{\arraystretch}{1.12}
\setlength{\tabcolsep}{3.5pt}
\caption{Matched statistical-encoder differences. Positive \(\Delta R^2_{\mathrm{all}}\) and positive Tail95 reductions favor the first model.}
\label{tab:direct_vs_raspl}
\begin{tabular}{lccc}
\hline
Comparison &
\(\Delta R^2_{\mathrm{all}}\) &
T95 reduction &
Wins \(R^2/\mathrm{T95}\) \\
\hline
RASPL vs.\ direct
& \(+0.733\)
& \(0.919\)
& \(9/9,\ 9/9\) \\
RASPL vs.\ formula feature
& \(+0.015\)
& \(0.061\)
& \(8/9,\ 8/9\) \\
Formula feature vs.\ direct
& \(+0.717\)
& \(0.858\)
& \(9/9,\ 9/9\) \\
\hline
\end{tabular}
\end{table}

The matched convolutional comparison shows the same gain over direct prediction, but a more nuanced tradeoff relative to the formula-feature baseline. Across the 15 matched regime--seed cells, RASPL-CNN-RAW+STATS exceeds Direct-CNN-RAW+STATS by \(0.3427\) in mean \(R^2_{\mathrm{all}}\), reduces Tail95 MAE by \(0.2482\), and achieves a paired DRS improvement of \(0.8950\). Relative to the formula-feature CNN, the average \(R^2_{\mathrm{all}}\) is slightly lower for RASPL (\(0.8026\) versus \(0.8060\)), but RASPL yields lower Tail95 MAE (\(0.2329\) versus \(0.2797\)), lower Tail95 underprediction (\(0.2015\) versus \(0.2412\)), and stronger paired degradation robustness. Thus, formula preservation does not uniformly maximize every individual metric, but it provides the more favorable robustness--tail-error tradeoff in the matched CNN setting.

\begin{figure*}[!t]
\centering

\begin{minipage}{\textwidth}
\centering
\small
\setlength{\tabcolsep}{4.5pt}
\renewcommand{\arraystretch}{1.12}

\captionof{table}{Contextual-encoder and window-size comparison over three degraded \(K\) regimes and seeds \(\{42,123,999\}\). DRS is normalized within the common four-model comparison pool.}
\label{tab:window_size_ablation}

\begin{tabular}{lcccccc}
\hline
RASPL variant &
\(R^2_{\mathrm{all}}\) \(\uparrow\) &
\(R^2_{\mathrm{pos}}\) \(\uparrow\) &
T95 MAE \(\downarrow\) &
T95 Under \(\downarrow\) &
DRS \(\uparrow\) &
Parameters \(\downarrow\) \\
\hline
MLP-STATS
& \textbf{0.834251}
& 0.870020
& 0.238474
& 0.253847
& 0.526799
& \textbf{3,458} \\
CNN-RAW+STATS, \(3\times3\)
& 0.827687
& \textbf{0.877491}
& \textbf{0.221776}
& \textbf{0.183496}
& \textbf{0.656387}
& 32,162 \\
CNN-RAW+STATS, \(5\times5\)
& 0.795051
& 0.847812
& 0.262103
& 0.229434
& 0.503745
& 32,162 \\
CNN-RAW+STATS, \(7\times7\)
& 0.806355
& 0.844393
& 0.258385
& 0.221979
& 0.508077
& 32,162 \\
\hline
\end{tabular}
\end{minipage}

\vspace{4pt}

\begin{minipage}{\textwidth}
\centering
\includegraphics[width=0.9\linewidth]
{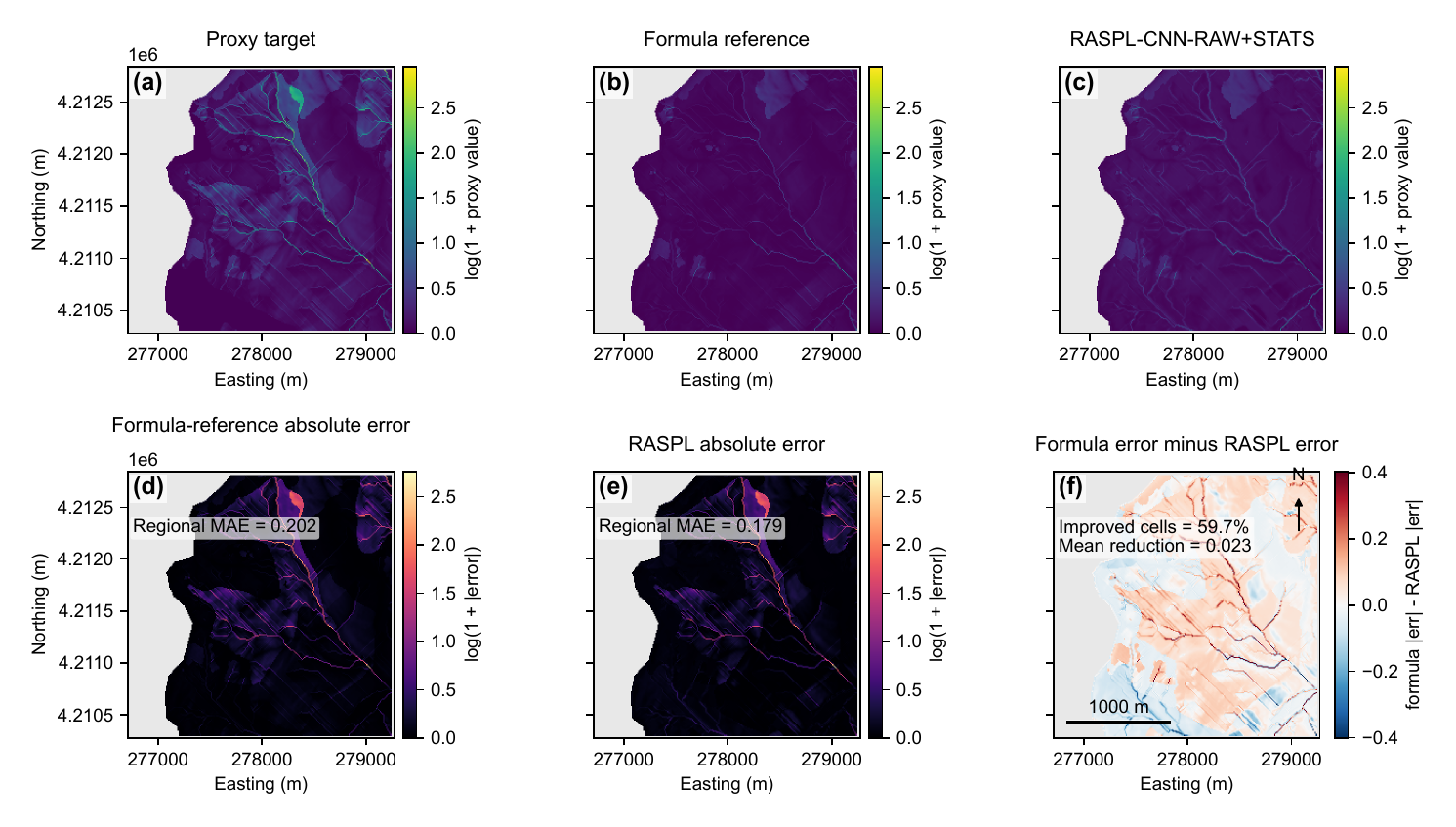}

\captionof{figure}{Spatial comparison under \(K_{\mathrm{missing}}\):
(a) proxy target, (b) formula-reference prediction using \(\bar{K}_{\mathrm{train}}\), (c) RASPL-CNN-RAW+STATS prediction,
(d)--(e) absolute errors, and (f) formula-reference error minus RASPL
error. Panels (a)--(c) use a shared \(\log(1+x)\) scale, panels
(d)--(e) use \(\log(1+|\mathrm{error}|)\), and panel (f) is symmetrically clipped at the \(99.5\) thpercentile; positive (warm) values favor RASPL, whereas negative (cool) values favor the formula reference.}
\label{fig:kmissing_spatial}
\end{minipage}

\end{figure*}

\textbf{Contextual Encoder Tradeoff.} Table~\ref{tab:window_size_ablation} compares the statistical RASPL encoder with convolutional encoders using \(3\times3\), \(5\times5\), and \(7\times7\) local windows. All results are macro-averaged over the same three degraded-factor regimes and three seeds. The reported DRS values are normalized within the common four-model comparison pool and therefore differ from the focused DRS values in Table~\ref{tab:main_results}.

RASPL-MLP-STATS attains the highest macro \(R^2_{\mathrm{all}}\) and uses only \(3{,}458\) parameters, compared with \(32{,}162\) for each CNN variant. Its mean historical training wall time is \(549.97\) seconds, compared with \(834.16\) seconds for the \(3\times3\) CNN. These results identify the statistical encoder as the most favorable average-accuracy and computational-efficiency operating point.

RASPL-CNN-RAW+STATS with a \(3\times3\) window instead achieves the highest \(R^2_{\mathrm{pos}}\), the lowest Tail95 MAE, the lowest Tail95 underprediction rate, and the highest DRS. The convolutional encoder therefore provides the stronger degradation- and tail-robustness operating point, despite its slightly lower macro \(R^2_{\mathrm{all}}\) and higher computational cost.

\textbf{Window-Size Ablation.} Increasing the CNN window beyond \(3\times3\) does not improve aggregate performance. Relative to \(3\times3\), the \(5\times5\) model decreases \(R^2_{\mathrm{all}}\) by \(0.032636\) and increases Tail95 MAE by \(0.040327\). The \(7\times7\) model decreases \(R^2_{\mathrm{all}}\) by \(0.021333\) and increases Tail95 MAE by \(0.036609\). Both larger windows also produce lower \(R^2_{\mathrm{pos}}\), higher Tail95 underprediction, and lower DRS.

The CNN parameter count remains fixed across window sizes, but the larger inputs increase computational cost. Relative to \(3\times3\), multiply--accumulate operations per sample increase by factors of \(2.70\) and \(5.25\) for \(5\times5\) and \(7\times7\), respectively. Because neither larger window improves accuracy or tail robustness, \(3\times3\) is retained as the convolutional configuration.

\textbf{Missing-Factor Stress Test.}
The \(K_{\mathrm{missing}}\) regime is evaluated separately as the strongest stress test and is not included in the shared main-ranking tables. Figure~\ref{fig:kmissing_spatial} visualizes the selected \(3\times3\) RASPL-CNN-RAW+STATS configuration on the non-overlapping \(256\times256\) candidate tile containing the largest number of Tail95 samples, with ties resolved by the smallest top-left raster index. This predefined rule avoids selecting the visualization according to appearance or model performance. On this tile, the formula reference yields regional MAE \(0.2023\) and Tail95 MAE \(1.1605\), whereas RASPL-CNN-RAW+STATS reduces these to \(0.1789\) and \(1.0574\), respectively. RASPL produces lower absolute error than the formula reference on \(59.70\%\) of valid cells, with a mean absolute-error reduction of \(0.0234\). The Tail95 underprediction rate nevertheless remains high for both models (\(1.0000\) for the formula reference and \(0.9732\) for RASPL), showing that complete removal of \(K\) remains a severe stress test even when average error improves.

\section{Conclusions and Future Work}

Using a RUSLE-derived soil-loss proxy, we showed that high complete-factor accuracy can reflect equation reconstruction rather than robustness to degraded factors. We introduced a diagnostic framework and RASPL, which retains the available formula estimate as an adaptively gated prediction anchor. Matched experiments show that formula-aware models outperform direct prediction and that explicit formula preservation yields a stronger degradation--tail-error tradeoff than treating the formula estimate as an ordinary feature. RASPL-MLP-STATS provides the best accuracy--efficiency tradeoff, whereas RASPL-CNN-RAW+STATS with a \(3\times3\) window provides the strongest degradation and tail robustness. Future work will test RASPL across regions and other factor-derived scientific tasks.

\bibliography{aaai2026}

@article{Benavidez2018RUSLEReview,
  author    = {Benavidez, R. and Jackson, B. and Maxwell, D. and Norton, K.},
  title     = {A review of the (Revised) Universal Soil Loss Equation (R/USLE): with a view to increasing its global applicability and improving soil loss estimates},
  journal   = {Hydrology and Earth System Sciences},
  volume    = {22},
  pages     = {6059--6086},
  year      = {2018},
  doi       = {10.5194/hess-22-6059-2018},
  publisher = {Copernicus Publications}
}

@article{Panagos2015,
  title   = {The new assessment of soil loss by water erosion in Europe},
  author  = {Panagos, Panos and Borrelli, Pasquale and Poesen, Jean and Ballabio, Carlo and Lugato, Emanuele and Meusburger, Katrin and Montanarella, Luca and Alewell, Christine},
  journal = {Environmental Science \& Policy},
  volume  = {54},
  pages   = {438--447},
  year    = {2015},
  doi     = {10.1016/j.envsci.2015.08.012}
}

@article{Li2023RUSLE,
  author  = {Li, Pingheng and Tariq, Aqil and Li, Qingting and Ghaffar, Bushra and Farhan, Muhammad and Jamil, Ahsan and Soufan, Walid and El Sabagh, Ayman and Freeshah, Mohamed},
  title   = {Soil Erosion Assessment by {RUSLE} Model Using Remote Sensing and {GIS} in an Arid Zone},
  journal = {International Journal of Digital Earth},
  year    = {2023},
  volume  = {16},
  number  = {1},
  pages   = {3105--3124},
  doi     = {10.1080/17538947.2023.2243916}
}

@article{Ge2023,
  author = {Yuankai Ge and Longlong Zhao and Jinsong Chen and Xiaoli Li and Hongzhong Li and Zhengxin Wang and Yanni Ren},
  title        = {Study on Soil Erosion Driving Forces by Using (R)USLE Framework and Machine Learning: A Case Study in Southwest China},
  journal      = {Land},
  volume       = {12},
  number       = {3},
  ARTICLE-NUMBER = {639},
  year         = {2023}
}

@Article{land13020174,
AUTHOR = {Samarinas, Nikiforos and Tsakiridis, Nikolaos L. and Kalopesa, Eleni and Zalidis, George C.},
TITLE = {Soil Loss Estimation by Water Erosion in Agricultural Areas Introducing Artificial Intelligence Geospatial Layers into the RUSLE Model},
JOURNAL = {Land},
VOLUME = {13},
YEAR = {2024},
NUMBER = {2},
ARTICLE-NUMBER = {174}
}

@Article{w17091351,
AUTHOR = {Hlal, Mohammed and El Monhim, Bilal and Chenal, Jérôme and Munyaka, Jean-Claude Baraka and Azmi, Rida and Sbai, Abdelkader and Cwick, Gary and Hichou, Badr Ben},
TITLE = {Application of Deep Learning and Geospatial Analysis in Soil Loss Risk in the Moulouya Watershed, Morocco},
JOURNAL = {Water},
VOLUME = {17},
YEAR = {2025},
NUMBER = {9},
ARTICLE-NUMBER = {1351}
}

@article{Abiye2025SoilErodibility,
  author  = {Abiye, Wudu and Dengiz, Orhan},
  title   = {Digital mapping of soil erodibility factor in response to land use change using machine learning models},
  journal = {Environmental Systems Research},
  year    = {2025},
  volume  = {14},
  doi     = {10.1186/s40068-025-00402-w}
}

@article{BOUAMRANE2024998,
author = {Asma Bouamrane and Hamouda Boutaghane and Ali Bouamrane and Noura Dahri and Habib Abida and Mohamed Saber and Sameh A. Kantoush and Tetsuya Sumi},
title = {Soil erosion susceptibility prediction using ensemble hybrid models with multicriteria decision-making analysis: Case study of the Medjerda basin, northern Africa},
journal = {International Journal of Sediment Research},
volume = {39},
number = {6},
pages = {998-1014},
year = {2024}
}

@ARTICLE{Gelete_2024,
author = {Tadele Bedo Gelete and Pernaidu Pasala and Nigus Gebremedhn Abay and Gezahegn Weldu Woldemariam and Kalid Hassen Yasin and Erana Kebede and Ibsa Aliyi},        
TITLE={Integrated machine learning and geospatial analysis enhanced gully erosion susceptibility modeling in the Erer watershed in Eastern Ethiopia},        
JOURNAL={Frontiers in Environmental Science},
VOLUME={Volume 12 - 2024},
YEAR={2024}
}

@article{MomeniDamaneh2023ErosionML,
  author   = {Momeni Damaneh, Javad and Safdari, Ali Akbar and Azarnejad, Nazanin and Ghorbani, Majid and Panahi, Fatemeh and Afzali, Sayed Fakhreddin and Loppi, Stefano},
  title   = {Modeling Soil Erosion Susceptibility Using Machine Learning Techniques: Rud-e-Faryab Basin, Iran},
  journal = {Land Degradation \& Development},
  volume  = {36},
  number  = {18},
  pages   = {6396--6409},
  year    = {2025},
  publisher = {Wiley}
}

@article{Tkeshelashvili2024Tsageri,
  author  = {Tkeshelashvili, N.},
  title   = {Empirical and Machine Learning Models for Soil Erosion Risk Assessment: A Case Study of Tsageri Municipality, Georgia},
  journal = {Journal of Geography, Environment and Earth Science International},
  volume  = {28},
  number  = {11},
  pages   = {148--162},
  year    = {2024},
  publisher = {Science Domain International}
}

@article{Olii2025SaddangSEV,
  author    = {Olii, Muhammad Ramdhan and Zailani Olii, Abdul Kadir and Olii, Aleks and Djau, Rahman Abdul and Mokoagow, Muhamad Alfaikar and Kironoto, Bambang Agus and Bachtiar, Bachtiar and Olii, Rizky Selly Nazarina and Pakaya, Ririn},
  title    = {Tree-based machine learning algorithms for soil erosion vulnerability (SEV) prediction in Saddang Watershed, south Sulawesi, Indonesia},
  journal  = {Journal of Water and Climate Change},
  volume   = {16},
  number   = {4},
  pages    = {1459--1476},
  year     = {2025},
  publisher = {IWA Publishing}
}

@article{Priyadharshini2025GeomorphicRisk,
 author = {Priyadharshini V.M. and Ghadah Aldehim and Noha Negm and S. Subathradevi},
  title    = {Integrating geospatial techniques and machine learning for assessing soil erosion and associated geomorphic risks},
  journal  = {Journal of South American Earth Sciences},
  volume   = {156},
  pages    = {105463},
  year     = {2025},
  publisher = {Elsevier}
}

@article{Islam2025GullyErosionML,
  author   = {Islam, Fakhrul and Bibi, Tahmina and Rehman, Nazir Ur and Davis, J. Brian and Aslam, Rana Waqar and Rebouh, Nazih Y. and Elmannai, Hela and Tariq, Aqil},
  title    = {GIS- and RS-based models for gully erosion susceptibility mapping using machine learning and remote sensing data},
  journal  = {Earth Surface Processes and Landforms},
  volume   = {50},
  number   = {15},
  year     = {2025},
  publisher = {Wiley}
}

@Article{s23218966,
  author = {Zhao, Shengyu and Tu, Kaiwen and Ye, Shutong and Tang, Hao and Hu, Yaocong and Xie, Chao},
  title = {Land Use and Land Cover Classification Meets Deep Learning: A Review},
  journal = {Sensors},
  volume = {23},
  year = {2023},
  number = {21},
  pages = {8966}
}

@INPROCEEDINGS{sentinelData,
  author={Karra, Krishna and Kontgis, Caitlin and Statman-Weil, Zoe and Mazzariello, Joseph C. and Mathis, Mark and Brumby, Steven P.},
  booktitle={2021 IEEE International Geoscience and Remote Sensing Symposium IGARSS}, 
  title={Global land use / land cover with Sentinel 2 and deep learning}, 
  year={2021},
  volume={},
  number={},
  pages={4704-4707}
}

@inproceedings{Cong2022SatMAE,
author = {Cong, Yezhen and Khanna, Samar and Meng, Chenlin and Liu, Patrick and Rozi, Erik and He, Yutong and Burke, Marshall and Lobell, David B. and Ermon, Stefano},
title = {SatMAE: pre-training transformers for temporal and multi-spectral satellite imagery},
year = {2022},
booktitle = {Proceedings of the 36th International Conference on Neural Information Processing Systems},
articleno = {15},
numpages = {15},
series = {NIPS '22}
}

@inproceedings{CROMA,
author = {Fuller, Anthony and Millard, Koreen and Green, James R.},
title = {CROMA: remote sensing representations with contrastive radar-optical masked autoencoders},
year = {2023},
booktitle = {Proceedings of the 37th International Conference on Neural Information Processing Systems},
articleno = {241},
numpages = {33},
series = {NIPS '23}
}

@inproceedings{GeoCLIP,
author = {Cepeda, Vicente Vivanco and Nayak, Gaurav Kumar and Shah, Mubarak},
title = {GeoCLIP: clip-inspired alignment between locations and images for effective worldwide geo-localization},
year = {2023},
booktitle = {Proceedings of the 37th International Conference on Neural Information Processing Systems},
articleno = {379},
numpages = {12},
series = {NIPS '23}
}

@article{raissi2019physics,
title = {Physics-informed neural networks: A deep learning framework for solving forward and inverse problems involving nonlinear partial differential equations},
journal = {Journal of Computational Physics},
volume = {378},
pages = {686-707},
year = {2019},
issn = {0021-9991},
doi = {https://doi.org/10.1016/j.jcp.2018.10.045},
url = {https://www.sciencedirect.com/science/article/pii/S0021999118307125},
author = {M. Raissi and P. Perdikaris and G.E. Karniadakis}
}

@article{willard2022integrating,
author = {Willard, Jared and Jia, Xiaowei and Xu, Shaoming and Steinbach, Michael and Kumar, Vipin},
title = {Integrating Scientific Knowledge with Machine Learning for Engineering and Environmental Systems},
year = {2022},
issue_date = {April 2023},
volume = {55},
number = {4},
issn = {0360-0300},
url = {https://doi.org/10.1145/3514228},
doi = {10.1145/3514228},
journal = {ACM Comput. Surv.},
month = {Nov.},
articleno = {66},
pages   = {1--37},
numpages = {37}
}

@article{karniadakis2021physics,
  title   = {Physics-informed machine learning},
  author  = {Karniadakis, George Em and Kevrekidis, Ioannis G. and Lu, Lu and Perdikaris, Paris and Wang, Sifan and Yang, Liu},
  journal = {Nature Reviews Physics},
  volume  = {3},
  number  = {6},
  pages   = {422--440},
  year    = {2021},
  doi     = {10.1038/s42254-021-00314-5}
}

@InProceedings{yoon2018gain,
  title = {{GAIN}: Missing Data Imputation using Generative Adversarial Nets},
  author =  {Yoon, Jinsung and Jordon, James and van der Schaar, Mihaela},
  booktitle = {Proceedings of the 35th International Conference on Machine Learning},
  pages = {5689--5698},
  year = 	{2018},
  volume = {80},
  month = {Jul.}
}

@inproceedings{hendrycks2019benchmarking,
  title     = {Benchmarking Neural Network Robustness to Common Corruptions and Perturbations},
  author    = {Hendrycks, Dan and Dietterich, Thomas},
  booktitle = {International Conference on Learning Representations},
  year      = {2019}
}

@article{shorten2019survey,
  title   = {A survey on Image Data Augmentation for Deep Learning},
  author  = {Shorten, Connor and Khoshgoftaar, Taghi M.},
  journal = {Journal of Big Data},
  volume  = {6},
  number  = {60},
  year    = {2019},
  doi     = {10.1186/s40537-019-0197-0}
}

@article{huber1964robust,
author = {Peter J. Huber},
title = {{Robust Estimation of a Location Parameter}},
volume = {35},
journal = {The Annals of Mathematical Statistics},
number = {1},
publisher = {Institute of Mathematical Statistics},
pages = {73 -- 101},
year = {1964},
doi = {10.1214/aoms/1177703732}
}

@article{Borrelli2017GlobalErosion,
  title   = {An assessment of the global impact of 21st century land use change on soil erosion},
  author  = {Borrelli, Pasquale and Robinson, David A. and Fleischer, Larissa R. and Lugato, Emanuele and Ballabio, Cristiano and Alewell, Christine and Meusburger, Katrin and Modugno, Sirio and Sch{\"u}tt, Brigitta and Ferro, Vito and Bagarello, Vincenzo and Van Oost, Kristof and Montanarella, Luca and Panagos, Panos},
  journal = {Nature Communications},
  volume  = {8},
  number  = {1},
  pages   = {2013},
  year    = {2017},
  doi     = {10.1038/s41467-017-02142-7}
}

@article{daly2002prism,
  author  = {Daly, Christopher and Gibson, Wayne and Taylor, George and Johnson, Gary and Pasteris, Paul},
  year    = {2002},
  title   = {A knowledge-based approach to the statistical mapping of climate},
  journal = {Climate Research},
  volume  = {22},
  pages   = {99--113}
}

@misc{usda_ssurgo,
  author       = {{U.S. Department of Agriculture, Natural Resources Conservation Service}},
  title        = {Soil Survey Geographic (SSURGO) Database},
  year         = {2023},
  howpublished = {\url{https://www.nrcs.usda.gov/resources/data-and-reports/soil-survey-geographic-database-ssurgo}},
  note         = {Accessed 2025-12-14}
}

@misc{usgs_3dep,
  author       = {{U.S. Geological Survey}},
  title        = {{USGS 1/3 Arc Second n39w108 20210312}},
  year         = {2021},
  howpublished = {U.S. Geological Survey, The National Map},
  url          = {PASTE_THE_EXACT_TILE_PAGE_URL_HERE},
  note         = {Accessed 2025-12-20}
}

@ARTICLE{Ahmet_Mediterranean, 
AUTHOR={İpek, Ahmet Faruk and Kahya, Ercan },       
TITLE={Spatiotemporal prioritization of soil erosion risk using the RUSLE model and CMIP6 projections under future climate scenarios in a Mediterranean watershed},      
JOURNAL={Frontiers in Environmental Science},    
VOLUME={14},
YEAR={2026},
DOI={10.3389/fenvs.2026.1760569} 
}

@article{Zeghmar2024,
  author  = {Zeghmar, Amer and Mokhtari, Elhadj and Marouf, Nadir},
  title   = {A Machine Learning Approach for {RUSLE}-Based Soil Erosion Modeling in the {Beni Haroun Dam Watershed}, Northeast Algeria},
  journal = {Earth Science Informatics},
  year    = {2024},
  volume  = {17},
  number  = {4},
  pages   = {2921--2936},
  doi     = {10.1007/s12145-024-01305-7},
  issn    = {1865-0481}
}

\end{document}